\documentclass[11pt]{article}
\usepackage{amsmath}
\usepackage{authblk}
\usepackage{bm}
\usepackage{booktabs}
\usepackage{caption}
\usepackage{enumitem}
\usepackage[margin=1in]{geometry}
\usepackage{graphicx}
\usepackage{mathrsfs}
\usepackage{multirow}
\usepackage{float}
\usepackage{subcaption}
\usepackage[colorlinks=true,citecolor=blue]{hyperref}
\usepackage[numbers,sort&compress]{natbib}

\usepackage{adjustbox}

\DeclareUnicodeCharacter{2009}{ }

\title{Automatic Classification of Circulating Blood Cell Clusters based on Multi-channel Flow Cytometry Imaging}

\author[1]{Suqiang Ma}
\author[2]{Subhadeep Sengupta}
\author[3]{Yao Lee}
\author[4]{Beikang Gu}
\author[5]{Xianyan Chen}
\author[4]{Xianqiao Wang}
\author[1]{Yang Liu}
\author[6,7]{Mengjia Xu}
\author[3,8]{Galit H. Frydman}
\author[1*]{He Li}
\affil[1]{School of Chemical, Materials, and Biomedical Engineering, University of Georgia, Athens, GA 30602}
\affil[2]{Institute for Artificial Intelligence, University of Georgia, Athens, GA 30602}
\affil[3]{Pathology Core, Unit for Laboratory Animal Medicine, University of Michigan, Ann Arbor, MI 48104}
\affil[4]{School of Environmental, Civil, Agricultural and Mechanical Engineering, College of
Engineering, University of Georgia, Athens, GA, 30602}
\affil[5]{College of
Public Health, University of Georgia, Department of Epidemiology \& Biostatistics, Athens, GA, 30602}
\affil[6]{Department of Data Science, Ying Wu College of Computing, New Jersey Institute of Technology, Newark, New Jersey, United States.}
\affil[7]{McGovern Institute for Brain Research, Massachusetts Institute of Technology, Cambridge, Massachusetts, United States.}
\affil[8]{Division of Trauma, Emergency Surgery, and Surgical Critical Care at the Massachusetts General Hospital, Boston, Massachusetts, United States.}
\affil[*]{he.li3@uga.edu}
\date{}

\begin{document}

\maketitle

\begin{abstract}
Circulating blood cell clusters (CCCs) containing red blood cells (RBCs), white blood cells (WBCs), and platelets are significant biomarkers linked to conditions like thrombosis, infection, and inflammation. Flow cytometry, paired with fluorescence staining, is commonly used to analyze these cell clusters, revealing cell morphology and protein profiles. While computational approaches based on machine learning have advanced the automatic analysis of single-cell flow cytometry images, there is a lack of effort to build tools to automatically analyze images containing CCCs. Unlike single cells, cell clusters often exhibit irregular shapes and sizes. In addition, these cell clusters often consist of heterogeneous cell types, which require multi-channel staining to identify the specific cell types within the clusters. This study introduces a new computational framework for analyzing CCC images and identifying cell types within clusters. Our framework uses a two-step analysis strategy. First, it categorizes images into cell cluster and non-cluster groups by fine-tuning the You Only Look Once(YOLOv11) model, which outperforms traditional convolutional neural networks (CNNs), Vision Transformers (ViT). Then, it identifies cell types by overlaying cluster contours with regions from multi-channel fluorescence stains, enhancing accuracy despite cell debris and staining artifacts. This approach achieved over 95\% accuracy in both cluster classification and phenotype identification. In summary, our automated framework effectively analyzes CCC images from flow cytometry, leveraging both bright-field and fluorescence data. Initially tested on blood cells, it holds potential for broader applications, such as analyzing immune and tumor cell clusters, supporting cellular research across various diseases.

\end{abstract}


\section{Introduction}

\textbf{Background and significance.}Coronavirus Disease 2019 (COVID-19), caused by SARS-CoV-2, is characterized by systemic hyperinflammation, cytokine storms, and an elevated risk of thrombotic complications~\cite{cardone2020lessons}. The immune-thrombotic state induced by COVID-19 is a critical aspect of its pathogenesis, with current evidence indicating it is more severe than that caused by other viral pneumonia, such as Influenza H1N1~\cite{porfidia2020venous,stals2021risk,bonaventura2021endothelial}. The recognized prothrombotic state associated with COVID-19 has spurred research into the pathology related to COVID-induced immunothrombosis. Potential mechanisms mediating COVID-related thrombosis include endothelial dysfunction, shifts in the procoagulant and antifibrinolytic plasma protein phenotype, platelet hyperactivity, and neutrophil extracellular traps (NETs)~\cite{furie1995molecular,frydman2017technical, swystun2016role,connors2020thromboinflammation}. It is likely that a combination of these factors contributes to the prothrombotic state of COVID-19, with the exact contributors varying based on the patient's disease state. For example, emerging clinical studies have shown that circulating cell clusters (CCCs), such as platelet-white blood cell (WBC) clusters, platelet-red blood cell (RBC) clusters, and WBC clusters, in the blood samples from patients with COVID-19 may serve as a nidus for immuno-thrombosis~\cite{Gallastegi2021,le2020neutrophil,do2020vivo,nicolai2020immunothrombotic}. These CCCs could also have an adverse impact on the physiological functions of microvasculature by directly blocking small blood vessels in the microvasculature. 

 In our recent study~\cite{dorken2023circulating}, we have systematically studied the correlation between the various types of CCCs and the clinical outcomes of patients with COVID-19. In this study, 46 blood samples from 37 COVID-19 patients and 12 samples from healthy controls were analyzed with imaging flow cytometry. The results of imaging flow cytometry were used to detect, differentiate, and quantify CCC phenotypes for healthy controls and patients with COVID-19. It is noted that in this study, conventional imaging flow cytometry data analysis was performed using ``manual gating'', as illustrated in Fig.~\ref{figure1}(A, I-II-IV)~\cite{dorken2021circulating,dorken2023circulating}, which pertains to several drawbacks, e.g., hard to reproduce, subjective and biased, and time-consuming particularly for large datasets. 
 
 {\bf Related work.} On the other hand, with the significant advances in graphics processing unit (GPU) computational power over the past decade, artificial intelligence techniques and computer-aided analysis have become essential tools for automatically extracting critical features from biomedical images across various modalities, aiding in disease diagnosis and prognosis~\cite{shen2017deep,haque2020deep,zhang2022aoslo}. Specifically, deep convolutional neural networks (DCNNs) have been extensively utilized for the efficient identification and classification of blood cells~\cite{qin2018fine,kadry2022automated,lin2018leukocyte,wang2020human,zhang2020hybrid}. Unlike traditional methods that rely on hand-crafted features, DCNNs offer a more powerful approach. The combination of convolutional layers and subsampling layers within CNNs serve as feature extractors, capable of learning spatial hierarchical features directly from input images, thereby enhancing performance and accuracy. So far, artificial intelligence (AI) models based on DCNNs have achieved good performance in many applications involving blood cell classification~\cite{su2014neural,tomari2014computer,zhu2022retracted,gavas2021deep,macawile2018white,habibzadeh2018automatic,sharma2019white,huang2019blood,singh2020blood,deng2021deep,xu2017deep,zhang2018rbc,zhang2020automated,parab2021red,cheuque2022efficient,ha2022fine}.


{\bf Knowledge Gap.} Despite the fact that AI models have been widely employed to perform blood cell classification, existing models were mostly designed to classify single blood cells and bio-particles without considering cell clusters containing heterogeneous cell types, which requires information from not only the bright-field images, but also multiple staining channels. In addition, in contrast to single cells, cell clusters often exhibit irregular shapes and varied sizes, depending on the number of cells and how they interact. Herein, we propose to develop a computational framework to perform automatic analysis and classification of different types of CCCs based on the outputs from brightfield images and multi-channel of protein staining.  This framework is built on the You Only Look Once-11 (YOLOv8)~\cite{Jocher_YOLO_by_Ultralytics_2023}, the latest iteration of the renowned real-time object detection and image segmentation model, which has gained increasing popularity in various domains, including image classification~\cite{yang2024research,palanivel2023art,ferdi2024quadratic,luong2023detection}. Specifically, \emph{Chen et al.} introduced NBCDC-YOLOv8, which enhanced YOLOv8 with SPD-Conv, MultiSEAM, and BiFPN modules for blood cell detection and classification, achieving $\sim$94.7\% mAP on the BCCD dataset~\cite{chen2024nbcdc}.  A white blood cell detection study applied YOLOv8 to identify leukemia subtypes from $\sim$3,629 images, reporting 95.1\% accuracy~\cite{duong2023yolov8wbc}.  An arXiv preprint demonstrates the use of YOLOv8 (and YOLOv11) in early detection of acute lymphoblastic leukemia, achieving $\sim$98.8\% accuracy~\cite{awad2024all}.

In this study, we will demonstrate the effectiveness of this framework by classifying various CCC phenotypes, including RBC clusters, platelet clusters, WBC clusters, and WBC+platelet clusters, from flow cytometry images of blood samples from patients with COVID-19.
As shown in Fig.~\ref{figure1}(B), 
In this study, we present a two-stage classification and phenotyping framework based on multi-channel imaging flow cytometry for the efficient and accurate detection and analysis of CCCs. Our principal contributions are summarized as follows:

 \noindent
1) \textbf{Two-stage pipeline with semi-automated annotation.} In the first stage, a lightweight YOLO11m-cl network performs binary classification of cluster versus non-cluster, achieving an accuracy of 95.03\%. In the second stage, YOLOv8-seg generates cluster masks, supplemented by a semi-automated annotation workflow using Meta’s Segment Anything Model to produce 372 high-quality masks. These masks are combined with HSV-based thresholding to extract CD61 and CD45 signals for rapid and precise phenotyping.\\
2) \textbf{Multi-channel fusion strategy with noise mitigation.} We integrate bright-field and CD61/CD45 fluorescence channels, employing HSV-based threshold segmentation and a 15\% overlap criterion to accurately distinguish platelet-only, leukocyte-only, and mixed clusters. This strategy effectively mitigates noise-induced signal overlap during image acquisition, ensuring robust phenotype discrimination.\\
3) \textbf{A semi-automated annotation workflow.} We adopt a semi-automated annotation workflow that leverages Meta-SAM to generate high-quality segmentation masks with minimal manual correction, enabling accurate phenotypic analysis with reduced labeling cost.\\
4) \textbf{Comprehensive benchmarking.} Using 5-fold cross-validation on COVID-19 and healthy samples, our pipeline achieved 95.03\% classification accuracy, 72.3\% mask mAP\@0.5:0.95, and 91.2\% phenotyping accuracy. Moreover, YOLOv8m-seg outperformed Ultralytics’ pretrained YOLOv9-seg (81.48\%) and YOLO11-seg (81.39\%) with a mask mAP of 85.61\%, while YOLO11-cls provided the optimal accuracy–speed trade-off compared to AlexNet, VGGNet, ResNet, DenseNet, EfficientNet, and ViT.

\section{Data and Methods}
This section can be roughly divided into the following
three parts: experimental dataset resource and processing,  
 cell cluster classification and the classification of the phenotype of CCCs.
 
\subsection{Dataset resource and processing}
\subsubsection{Dataset resource}

\begin{figure}[H]
\centering
\includegraphics[width=0.7\linewidth]{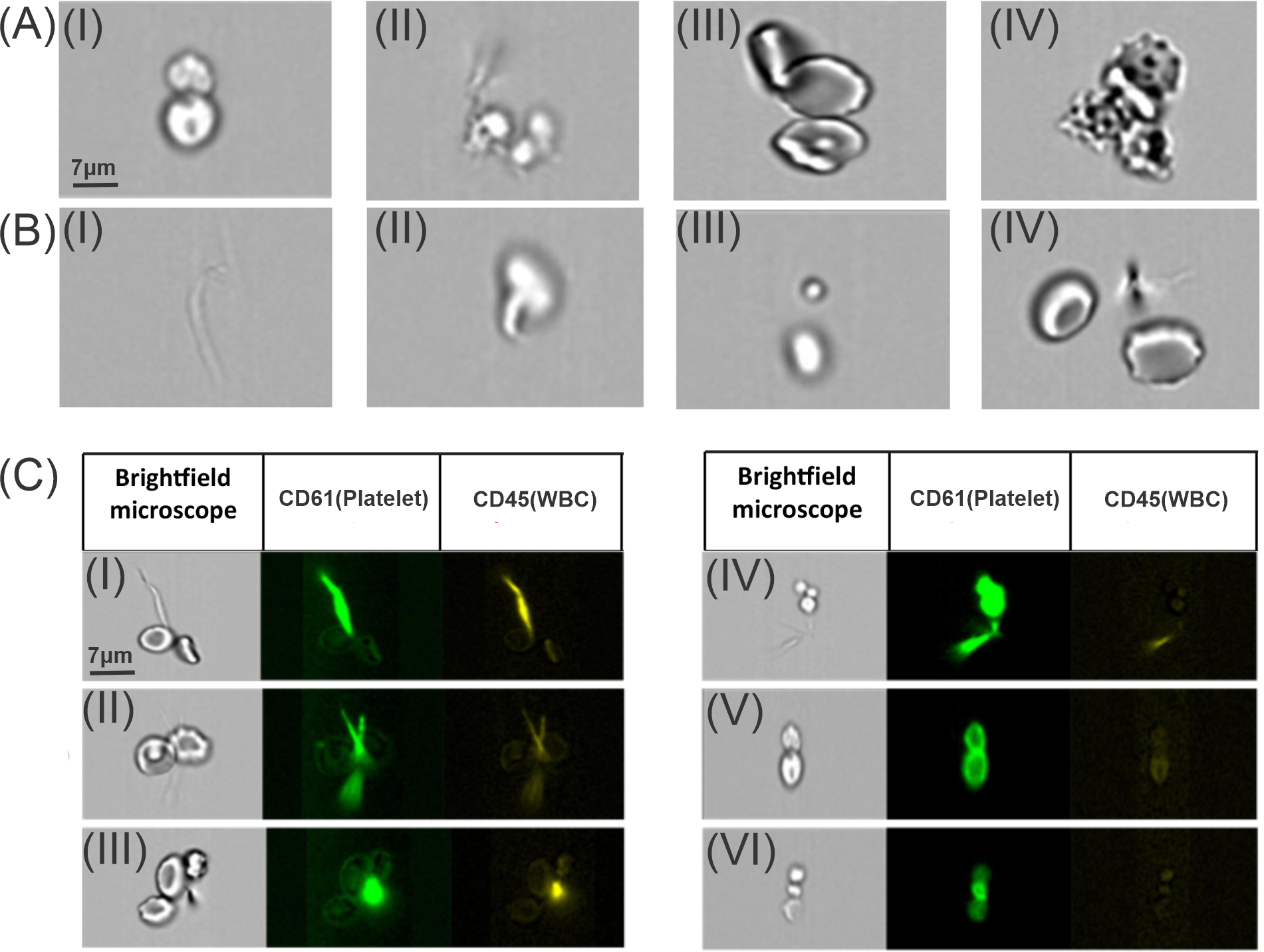}
\caption{(A) Examples of brightfield images of cell clusters. (B) Examples of brightfield images containing no cell clusters. (B,I) no cells; (B,II) single cells; (B,III-IV) multiple separated cells. (C) Images containing cell clusters with fluorescent staining. The left column displays the brightfield images. The color of fluorescence-stained regions is used to identify specific cell cluster types, with green fluorescence-stained regions corresponding to platelets and yellow fluorescence-stained regions indicating WBCs. (C,I-III) Images are featured with staining artifacts. (C,IV-VI) Images with normal staining. }
\label{figure2}
\end{figure}

The dataset comprises 1,568 grayscale microscopic images derived from 46 blood samples, collected from 37 COVID-19 patients and 12 healthy controls at Massachusetts General Hospital (Boston, MA), under IRB approval (protocol \#:2020P001364). All samples were acquired using multi-channel imaging flow cytometry, including both brightfield and fluorescence modalities. To ensure data quality, experienced hematopathologists manually reviewed and curated the raw image pool, systematically filtering out invalid samples—such as defocused frames, empty backgrounds, and debris-contaminated fields—while retaining only diagnostically meaningful images. This rigorous selection process ensured that the final dataset encompasses a comprehensive spectrum of morphologies, including tight and loose clusters, as well as diverse non-cluster states (single cells, multiple separated cells, and blank fields). Representative examples of this diversity are shown in Fig.2~\ref{figure2}, which includes typical brightfield cluster (A) and non-cluster (B) samples, as well as fluorescent overlays (C) highlighting platelet (CD61, green) and white blood cell (CD45, yellow) staining patterns under various conditions, including cases with minor fluorescence artifacts and well-resolved phenotypic signals. This carefully curated dataset provides a robust foundation for training and evaluating automated models for CCC classification and phenotyping under real-world imaging conditions.

\subsubsection{Dataset processing}
We constructed an annotated dataset of 1,568 grayscale microscopic images, each manually labeled as either ''cluster'' (indicating aggregated cells) or ''non-cluster'' (indicating dispersed, non-aggregated cells). The raw images exhibited considerable variation in spatial resolution and aspect ratio. To ensure compatibility with ImageNet-pretrained convolutional architectures, we applied a standardized preprocessing pipeline. First, each image was converted to a three-channel RGB format. Then, to preserve aspect ratio while achieving square dimensions, we computed the maximum of the image’s height and width, and symmetrically padded the shorter side with gray pixels (RGB = [173, 173, 173]). The padded images were resized to \(224 \times 224\) pixels using bilinear interpolation and saved in RGB format.

To enhance data diversity and improve model generalization in the cell cluster classification task, we implemented data augmentation strategies tailored to the underlying architecture type. For CNN-based models, we adopted a fixed-rate multi-augmentation scheme wherein each training image was dynamically augmented five times per epoch. This approach effectively simulated an offline fivefold expansion of the training set while maintaining storage efficiency through on-the-fly augmentation. The augmentation pipeline included random resized cropping (scale: 0.8–1.0), full 360-degree rotation, horizontal and vertical flipping, color jittering (brightness, contrast, saturation, and hue), and Gaussian blurring. For YOLO-based models, we employed the default online augmentation pipeline provided by the Ultralytics framework. This pipeline applies a diverse set of transformations stochastically during training, including random affine warping, HSV jitter, flipping, mosaic composition, and MixUp. While CNNs use deterministic multi-instance augmentation and YOLO models apply stochastic online transformations, the resulting level of visual diversity was comparable. This allowed for a fair and balanced evaluation across model architectures.

To ensure robust generalization and mitigate overfitting, we adopted a five-fold cross-validation protocol. The entire dataset was partitioned into five approximately equal folds. In each iteration, four folds were used for training (with augmentation) and one for validation. This procedure was repeated five times so that each image served exactly once as a validation sample. The data distribution across folds, including augmented training sets, is summarized in Table~\ref{tab:dataset_splits_aug}.

\begin{table}[htbp]
\centering
\caption{Dataset splits across five folds, with augmentation counts (×5)}
\label{tab:dataset_splits_aug}
\begin{tabular}{c|cc|cc|cc}
\toprule
\multirow{2}{*}{Fold} & \multicolumn{2}{c|}{Train} & \multicolumn{2}{c|}{Augmented\_Train (×5)} & \multicolumn{2}{c}{Test} \\
                      & Cluster & Non-cluster & Cluster & Non-cluster & Cluster & Non-cluster \\
\midrule
0 & 682 & 572 & 3410 & 2860 & 173 & 141 \\
1 & 681 & 573 & 3405 & 2865 & 174 & 140 \\
2 & 679 & 575 & 3395 & 2875 & 176 & 138 \\
3 & 682 & 573 & 3410 & 2865 & 173 & 140 \\
4 & 696 & 559 & 3480 & 2795 & 159 & 154 \\
\bottomrule
\end{tabular}
\end{table}

\subsection{The method overview}

\begin{figure}[H]
\centering
\includegraphics[width=0.8\linewidth]{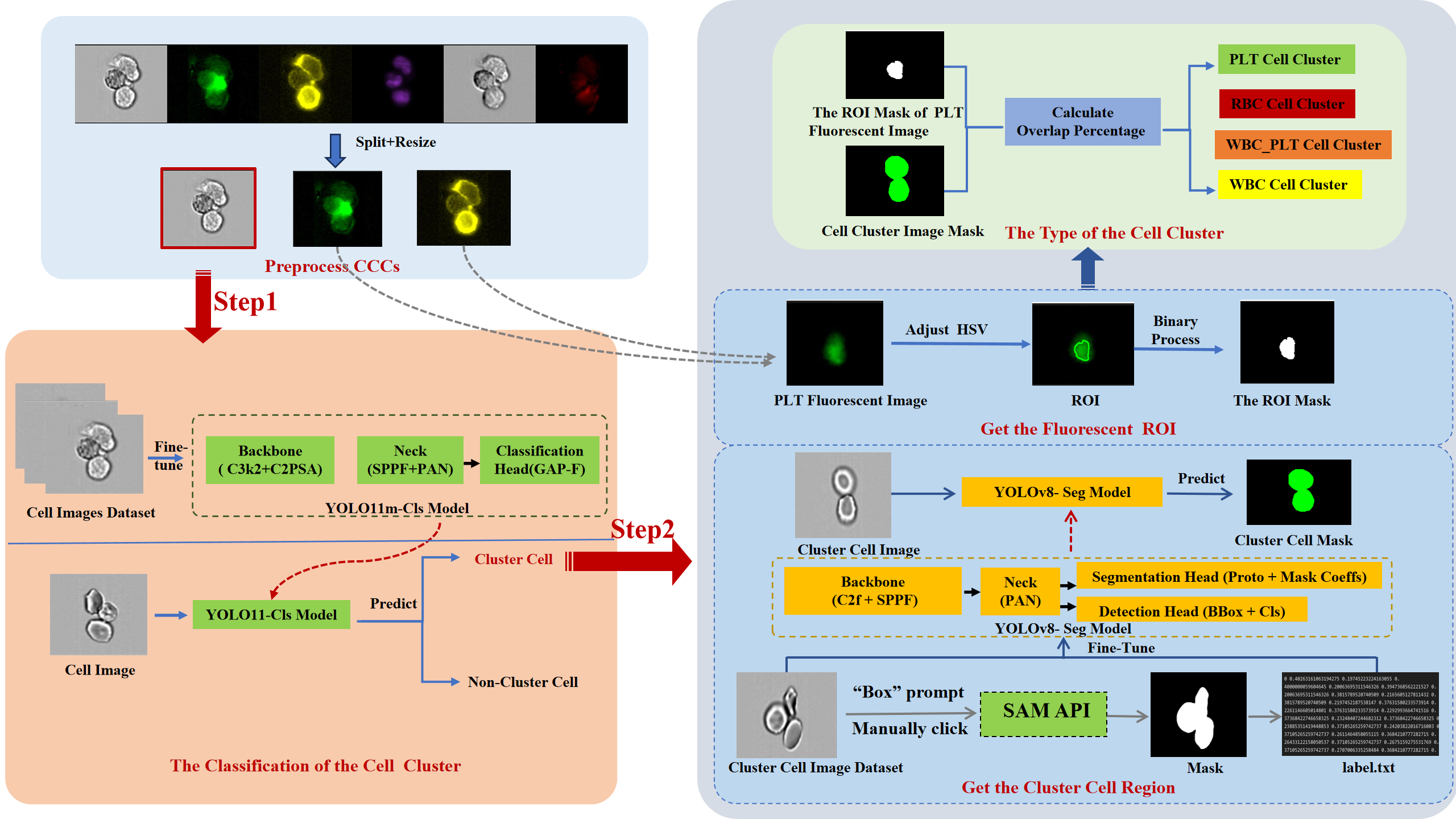}
\caption{The proposed framework automatically classifies CCCs using multi-channel flow cytometry imaging. The classification is performed through two sequential steps: I) We first filter out all the images that do not contain CCCs; II) We identify the cell types within the CCCs based on fluorescence-stained images. MetaSAM~\cite{kirillov2023segment} is applied for segmentation
mask generation, significantly reducing manual labeling effort while maintaining mask quality
over 372 cluster images.}
\label{figure1}
\end{figure}

In this study, we propose a fully automated, two-stage framework for the classification and subtyping of circulating cell clusters (CCCs) in multi-channel flow cytometry images. The framework is designed to systematically identify whether an image contains a cell cluster, and, if so, further determine its cellular composition based on morphological and fluorescence cues. As illustrated in Figure~\ref{figure1}, the pipeline consists of the following sequential steps:

\textbf{Step 1 – Cluster Cell Classification:}  
All microscopic images are first preprocessed through resizing, channel splitting, and standardization. A YOLO11-based classifier is then fine-tuned to discriminate between images that contain CCCs and those that do not. This step serves as an automated filtering mechanism, reducing downstream computational load and improving specificity in cluster-level analysis.

\textbf{Step 2 – Cell Type Identification within Clusters:}  
For images identified as containing CCCs, we further determine the biological subtype of the cluster (e.g., RBC cluster, PLT cluster, or WBC-PLT hybrid cluster). To this end, a YOLOv8-based segmentation model is fine-tuned to isolate the region of interest (ROI) from fluorescent channels. The fluorescent signal is then processed using color space transformation, binarization, and morphological filtering to generate ROI masks. The overlap between the masks and predefined fluorescence markers is quantitatively analyzed to assign the final cluster subtype.  In this stage, we leverage semi-automatic ROI
labeling tools, Segment Anything Model (SAM)~\cite{kirillov2023segment}, for annotation initialization during training. Specifically, the bounding boxes around cell clusters are manually annotated by trained users using a lightweight GUI. These boxes serve as prompts for SAM, which then generates binary segmentation masks. This approach is chosen to reduce manual labor while ensuring high-quality and consistent annotations. To reduce potential annotation variability, all bounding boxes are reviewed by at least one other annotator, and inconsistent results are excluded or re-annotated.

To ensure clarity and reproducibility, the following subsections describe the technical details of each component in this two-step framework.

\subsubsection{Cluster Cell Classification}

The goal of our study is to distinguish images containing circulating cell clusters (CCCs) from those showing single or non‐attached cells, based on multi‐channel imaging flow cytometry. This task is complicated by indistinct cluster boundaries, cell overlap, and imaging artifacts such as noise. Initial attempts using traditional OpenCV edge‐detection techniques proved insufficient due to the high variability and noise in the samples. To address these challenges, we assembled and manually annotated a dataset of 1,568 blood‐cell images from multiple patient samples, evenly divided into “cluster” and “non‐cluster” classes and split into training and testing sets at a 4:1 ratio.

We then conducted a systematic evaluation of several deep‐learning classifiers, each initialized with publicly available ImageNet‐pretrained weights and fine‐tuned under identical hyperparameter settings. Our benchmark included classic convolutional architectures—ResNet18 and ResNet34, which leverage residual identity‐mappings to ease optimization of deep networks~\cite{he2016deep}; DenseNet121, featuring dense connectivity for feature reuse and improved gradient flow~\cite{huang2017densely}; AlexNet, the pioneering deep CNN that popularized ReLU activations and dropout~\cite{krizhevsky2012imagenet}; and VGG16, built from uniform \(3\times3\) convolutions~\cite{simonyan2014very}—as well as modern alternatives: Vision Transformer (ViT‐B/16), which applies multi‐head self‐attention to image patches~\cite{dosovitskiy2020image}; and EfficientNet‐B0, employing compound scaling of depth, width, and resolution for efficiency~\cite{tan2019efficientnet}. 

In addition, we benchmarked the classification heads of the Ultralytics YOLO family. We selected YOLOv8 and YOLO11—each available in nano (n), small (s), medium (m), large (l), and extra‐large (x) sizes—because they are the only YOLO releases with publicly distributed ImageNet‐pretrained classification weights~\cite{jocher2023yolov8}. This enabled a fair comparison of accuracy and inference speed without the need to train from scratch.

Each model was fine‐tuned on the CCC dataset across five independent trials; mean classification accuracy and the distribution of misclassified samples were used to assess performance and robustness (see Fig.~\ref{figure2}). Ultimately, the YOLO11m‐cls variant demonstrated the best balance of high accuracy and real‐time processing capability and was selected for our primary binary classification stage.

{\bf YOLO11m-cls Model Architecture.} The YOLO11m-cls model is the medium-sized classification variant in the YOLO11 family (10.4 M parameters, $\approx$4.5 ms inference time), built on the standard backbone-neck-head paradigm with three key innovations:
\begin{enumerate}[leftmargin=2em,label=\arabic*)]
  \item \textbf{Backbone.}  
    Feature extraction is performed by a series of C3k2 modules---Cross Stage Partial blocks using compact $2\times2$ convolutions for parameter efficiency---and C2PSA units, which run a parallel spatial-attention branch alongside standard convolutions to emphasize salient image regions.
  \item \textbf{Neck.}  
    Multi‐scale feature aggregation is handled by a streamlined SPPF (Spatial Pyramid Pooling—Fast) layer, which applies a single pooling kernel at multiple dilation rates, followed by a lightweight PAN (Path Aggregation Network) that fuses high‐ and low‐resolution features for both coarse and fine details.
  
  \item \textbf{Head.}  
    A pure classification head replaces the detection head: a \(1\times1\) convolution projects the fused feature map to 512 channels; global average pooling reduces it to a 512-dimensional descriptor; and a fully connected layer (512 → 2) plus softmax produces the two-class (cluster vs.\ non-cluster) probabilities. In total, the network comprises approximately 120 convolutional layers, with the classification head adding only about 0.1 M parameters.
\end{enumerate}

{\bf Model Training.}   
YOLO11m‐cls is fine-tuned from ImageNet weights using SGD with momentum, weight decay, and a cosine-annealing learning-rate schedule. A combined cross-entropy and focal loss addresses class imbalance, improves robustness, and handles class imbalance. All layers are fine‐tuned from ImageNet weights using SGD with momentum (0.9), weight decay \(5\times10^{-4}\), an initial learning rate of 0.01 (linear warmup over 3 epochs), and cosine annealing over 100 epochs. This configuration enables YOLO11m‐cls to achieve superior accuracy (95.03\%) and real‐time inference speed, making it ideal for high‐throughput CCC screening.


{\bf Model evaluation.} 
To evaluate the performance of the classification model, we conduct 5-fold cross-validation, where the dataset is split into 5 folds. The model is trained on 4 folds and validated on the remaining fold. For each trial, we record classification accuracy, precision, recall, and F1 score. The average values from these 5 trials are calculated and used as the final results for each evaluation metric. This cross-validation approach helps mitigate the effects of randomness, ensuring the stability and reliability of the outcomes.

 The four metrics are computed using the following quantities. Here, we consider that cell clusters are positive samples and non-cell clusters are negative samples. 1. TP: true positives, the number of correctly classified cell clusters. 2. FP: false positives, the number of non-cell clusters classified as cell clusters. 3. TN: true negatives, the number of non-cell-cluster correctly classified. 4. FN: false negatives, the number of cell clusters classified as non-cell clusters. The accuracy, precision, recall, and F1 score are computed as follows:

\begin{align}
\mathrm{Accuracy} &= \frac{TP + TN}{TP + TN + FP + FN}\tag{1}\,,\\
\mathrm{Precision} &= \frac{TP}{TP + FP}\tag{2}\,,\\
\mathrm{Recall} &= \frac{TP}{TP + FN}\tag{3}\,,\\
F_1 &= 2 \times \frac{\mathrm{Precision}\times \mathrm{Recall}}{\mathrm{Precision} + \mathrm{Recall}}\tag{4}\,.
\end{align}

 Accuracy represents the proportion of correctly classified samples, including both positive and negative classes. It indicates the model's overall correctness across all predictions and can be calculated using Eq.(1). Precision measures the accuracy of positive predictions, representing the proportion of true positives among all positive predictions. It evaluates the model’s ability to avoid incorrectly classifying negative samples as positive (FP) and can be computed using Eq.(2). Recall assesses the model's ability to correctly identify positive samples, indicating the proportion of actual positive samples that are correctly predicted as positive. It reflects the model’s capacity to reduce missed classifications (FN) and can be calculated using Eq.(3). The F1 Score is the harmonic mean of Precision and Recall, providing a balanced evaluation of the model’s performance in handling both FP and FN. This metric is especially useful when dealing with imbalanced datasets and can be computed using Eq.(4).

For step 2 of the segmentation model, we evaluate performance using precision, recall, and Intersection over Union (IoU), including metrics like mAP@50 and mAP@50-95. The implementation is done in Python on a system equipped with two NVIDIA GeForce RTX 3090 GPUs, each with 24GB of memory. The PyTorch library was used as the primary framework, and the experiments were conducted on an Ubuntu 20.04.1 system.

\subsection{Classification of the phenotype of CCCs.}

For images containing cell clusters, we further classify various CCC phenotypes, including RBC clusters, platelet clusters, WBC clusters, and WBC+platelet clusters. Here, fluorescence staining is utilized to identify the specific types constituting the cell clusters. For instance, the presence of green color in the CD61 channel indicates that the CCCs contain platelets, while the presence of yellow color in the CD45 channel indicates that the CCCs contain WBCs. However, in some cases, as illustrated in Figs.~\ref{figure2}(C, I-III), the fluorescence-stained region is primarily located outside the CCCs, which is likely induced by the presence of the cell debris. These artifacts that occurred during the staining could lead to misclassification. Conversely, Figs.~\ref{figure2}(C, IV-VI) represent examples of normal staining. In Fig.~\ref{figure2}(C, IV), although there is some extent of artifacts in the staining, the green fluorescence entirely covers the original cell cluster region, suggesting that the cluster is composed of platelets. Fig.~\ref{figure2}(C, V) shows the normal staining, where the green fluorescence completely covers the original cell cluster region without any artifacts, indicating the presence of a platelet cell cluster. In Fig.~\ref{figure2}(C, VI), the fluorescence intensity covers only part of the original cell cluster, implying that this cell cluster contains other blood cell types in addition to platelets. 

Due to the presence of staining artifacts, as exemplified in Figs.~\ref{figure2}(C, I-III), accurate identification of cell cluster types requires a precise evaluation of the association between the stained area and the original cell cluster's boundary. If the stained areas in the images do not overlap with the cell cluster's contour, these images are excluded from the classification of cell cluster types. On the other hand, when the region of interest's fluorescent intensity lies within the cell cluster's contour, it is considered a valid staining result that can be used for subsequent classification.  In some cases, as shown in Fig.~\ref{figure2}(C, IV)(d), where the stained area overlaps with the cell cluster area but also goes beyond the cluster area, we continue with the subsequent classification. Moreover, a cell cluster may contain RBCs that are not stained.  As a result, the region of staining and the cell cluster area only partially overlap.

\subsubsection{Delineating cell cluster contours on brightfield images}

Given the complex, irregular morphology of cell clusters, we utilized the YOLOv8 model~\cite{Jocher_YOLO_by_Ultralytics_2023} for cell contour detection. The first step in training the YOLOv8 model involves generating label files that record both the cell cluster types and the coordinates of their contour points.  Herein, to accelerate the segmentation of cell clusters, we adopted a semi-automatic annotation workflow based on Meta’s Segment Anything Model (SAM)~\cite{kirillov2023segment}. Each input image was first displayed using a custom annotation interface. The region of interest (ROI) was manually selected by drawing a bounding box around the target cell area. The selected coordinates were then sent as a prompt to the SAM model API. Upon receiving the request, SAM produced a binary mask corresponding to the selected object. The generated mask was automatically overlaid on the original image for validation and saved to disk for downstream training. Importantly, annotators were instructed to draw tight bounding boxes that enclose only the cell regions while avoiding surrounding noise or irrelevant areas, to ensure accurate overlap calculation in subsequent analysis. 


In this task, a dataset comprising 372 images of various CCC types is partitioned into training and test sets, adhering to a 4:1 ratio for conducting the segmentation experiments. The SAM approach~\cite{kirillov2023segment} is employed to generate mask images by manually annotating the cell clusters using bounding box prompts.  The resulting mask images are preserved. As depicted in Fig.~\ref{figure1}(B, step2), the YOLOv8 label format is adapted to produce label text files that document the cluster types and contour coordinates. The original images, along with their corresponding label files, are compiled into a cohesive dataset for training the YOLOv8 segmentation model. When the model processes an image of a cell cluster, it generates a corresponding mask. The loss function of the YOLOv8 model includes box loss (measuring localization error for bounding boxes), segmentation loss (assessing the accuracy of segmented cell areas), and distribution focal loss (improving the precision of bounding box regression). The main workflow of our cell cluster segmentation method is shown in Fig.~\ref{figure1}(B, step 2).

\begin{figure}
\centering
\includegraphics[width=0.8\linewidth]{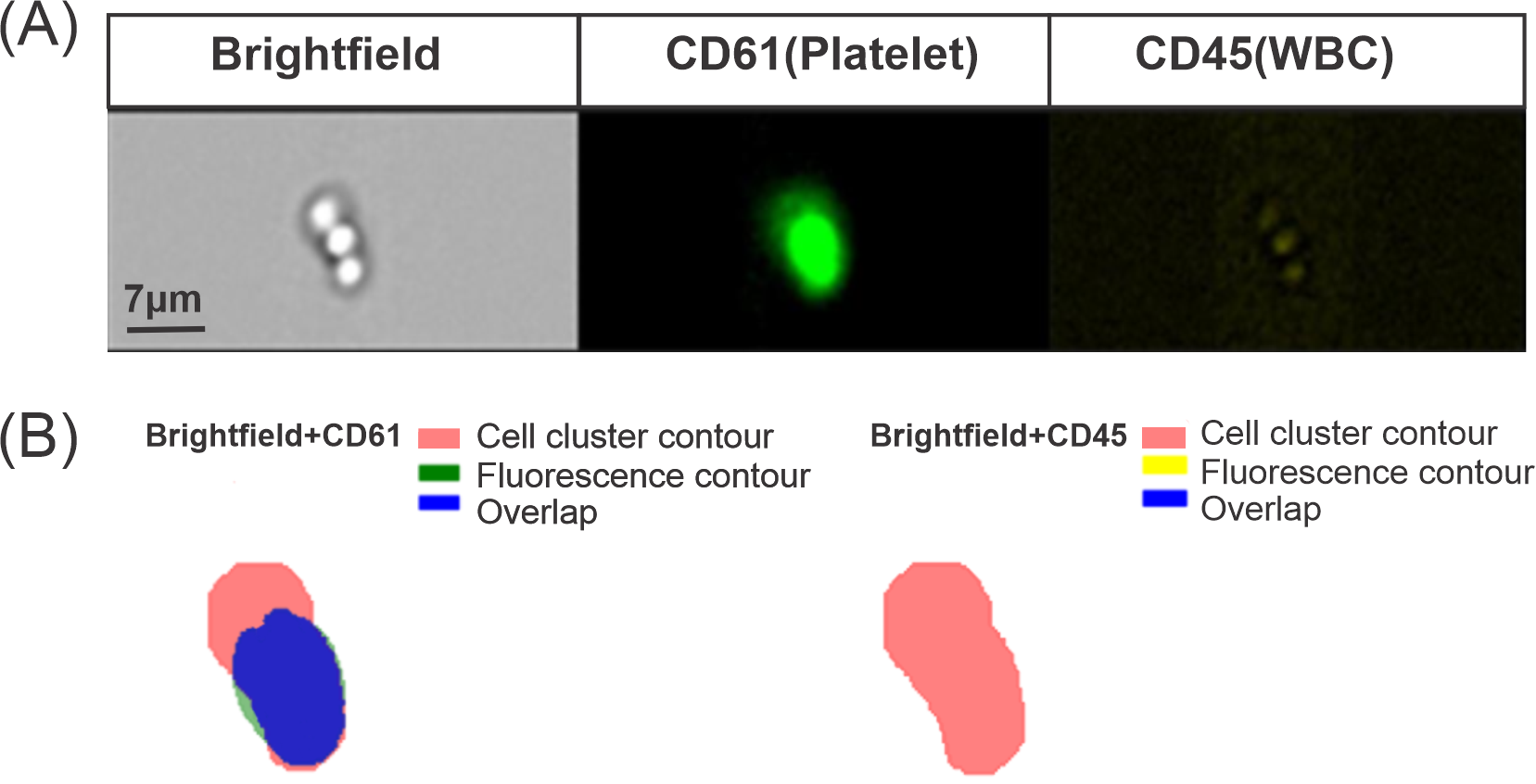}
\caption{Identify the cell cluster type by evaluating the overlap of the area of the cell cluster and fluorescence-stained regions. (A) The example of multi-channel flow cytometry images of blood cell clusters from one patient with COVID-19. (B) Identifying the cell types within the clusters by overlapping the fluorescent-stained region on the cell cluster mask extracted from brightfield images. Left: Overlaps are detected between the cell cluster mask and fluorescent-stained for CD61, indicating the presence of platelets within the cell clusters. Right: No overlapping is detected between the cell cluster mask and fluorescent staining for CD45, indicating the absence of WBCs within the cell clusters.}
\label{figure3}
\end{figure}

\subsubsection{Identification of colored regions in fluorescence-stained images. }

Next, we determine the type of CCCs based on the types of cells present within a cell cluster using information from fluorescent staining images, including the CD61 and CD45 channels. For the CD61 channel, if an image containing cell clusters shows green fluorescence above a certain brightness threshold, and the fluorescence-stained region overlaps with the contour of the cell cluster, the CCC is considered to contain platelets. Similarly, the yellow fluorescence in the CD45 channel is evaluated under the same conditions, and if met, the cell cluster is considered to contain WBCs.

By assessing the brightness of the fluorescence-stained region, we can delineate its boundary. To identify and extract the fluorescence-stained region, we employ a color thresholding technique. Specifically, we utilize the HSV (Hue, Saturation, Value) color space to define the range of fluorescent colors and convert the image into a binary mask through thresholding operations. This process involves the following steps:
\begin{itemize}
    \item Defining HSV Range: The thresholding process involves several key steps to isolate specific regions based on their color values in the HSV color space. First, we define the lower and upper limits for the HSV values. For the fluorescent green region, the HSV thresholds are set to [35, 100, X] for the lower limit and [85, 255, 255] for the upper limit. Similarly, for the fluorescent yellow region, the thresholds are defined as [20, 100, X] and [40, 255, 255]. These thresholds correspond to the Hue (H), Saturation (S), and Value (V) ranges, with ``X'' representing the lower threshold for fluorescence brightness. This value is critical in determining the contours of the fluorescence-stained regions and will be fine-tuned through further experiments to optimize detection accuracy.
    
    \item The hue range of 35 to 85 typically covers green and adjacent colors (from yellow-green to cyan-green). This range is employed in detecting and filtering regions representing platelets through green color. The hue range of 20 to 40 typically covers yellow and adjacent colors (from orange-yellow to orange-yellow). This range is employed to detect and filter regions representing WBCs. Since the fluorescence-stained region has only one color, in order to prevent interference from noise samples, we set the value of this region to a larger value within a reasonable range.  The saturation range of 100 to 255 ensures that only pixels with relatively high color saturation are selected. Low saturation values (close to 0) result in colors that appear grayish, whereas high saturation values (close to 255) lead to more vibrant colors.  The variable X represents the lower threshold for fluorescence brightness. The value 255 is the maximum brightness level, corresponding to fully bright pixels, typically pure white or nearly white regions. Selecting pixels with a brightness between X and 255 allows for the filtering of relatively bright regions in the cell cluster images while ignoring darker parts. This approach helps accurately identify the fluorescence-stained region and exclude noise, facilitating precise determinations about the cell cluster type.
    
    \item For each image, using a threshold-based color segmentation method, we perform image loading and threshold processing to identify regions that may contain platelets or WBCs. The first step involves extracting regions from the input image that meet specific color conditions. By converting the image to HSV values and applying thresholding techniques to create a binary mask, we can effectively identify the fluorescence-stained regions within the cell clusters. Under the predefined HSV mode, we search for pixels that fall within the specified HSV range. This function returns a binary mask where the regions of interest are marked as 1 (displayed as white), and non-interest areas are marked as 0 (displayed as black). In this manner, we process the image with green fluorescence and the image with yellow fluorescence to examine their overlapping with the contour of the cell clusters.
\end{itemize}

\subsubsection{Determining the phenotype of cell cluster based on combined brightfield images and fluorescence-stained images}
To determine the presence of a specific cell cluster type, we analyze the spatial interaction between the green and yellow fluorescence-stained regions and the original cell cluster contour. If there is overlap between the stained region and the original cell cluster region, and the area of overlap reaches a certain threshold, we infer the presence of a specific cell type (e.g., WBC, platelet). We implement an overlap detection method to assess this alignment, where overlapping regions between fluorescence and brightfield masks serve as indicators of specific cell types within a cluster.

As illustrated in Fig.~\ref{figure3}(B), we segment the multi-channel flow cytometry image into images for individual channels. The brightfield image, showing the shape of the cell cluster, is segmented using the YOLOv8 model to obtain a mask image. The second and third fluorescence-stained images undergo color thresholding to extract their respective regions of fluorescence staining. As shown in Fig.~\ref{figure3}(B), we then evaluate the overlap between the fluorescence-stained regions and the cell cluster regions. To quantitatively assess the spatial overlap between two binary masks, we conduct the following steps. First, the area of the overlap region is determined by counting the total number of non-zero pixels in the overlap mask. Next, the total area of the first binary mask is calculated by summing its non-zero pixels. Finally, the area of the overlapped region is divided by the total area of the mask of the brightfield images to obtain the overlapping percentage. This percentage measures the similarity or degree of overlap between the two masks. 

For the classification of cell cluster phenotype into  RBC clusters, platelet clusters, WBC clusters, and WBC+platelet clusters, we use accuracy as the final metric to evaluate the model's performance.

\section{Results}
\subsection{Results of classifying cell cluster and non-cell cluster groups.}

We conducted a comprehensive classification experiment using five-fold cross-validation on our dataset, where all models were trained and evaluated under a consistent protocol. The evaluation covered a broad spectrum of representative classification architectures, including classical convolutional neural networks (AlexNet, VGG16, ResNet18/34, DenseNet121, EfficientNet-B0), transformer-based models (ViT-B/16), and the Ultralytics YOLOv8 and YOLO11 classification families. For YOLO-based models, we systematically benchmarked five size variants—nano (n), small (s), medium (m), large (l), and extra-large (x)—ensuring a fair comparison across model scales.

\begin{figure}[H]
\centering
\includegraphics[width=0.6\linewidth]{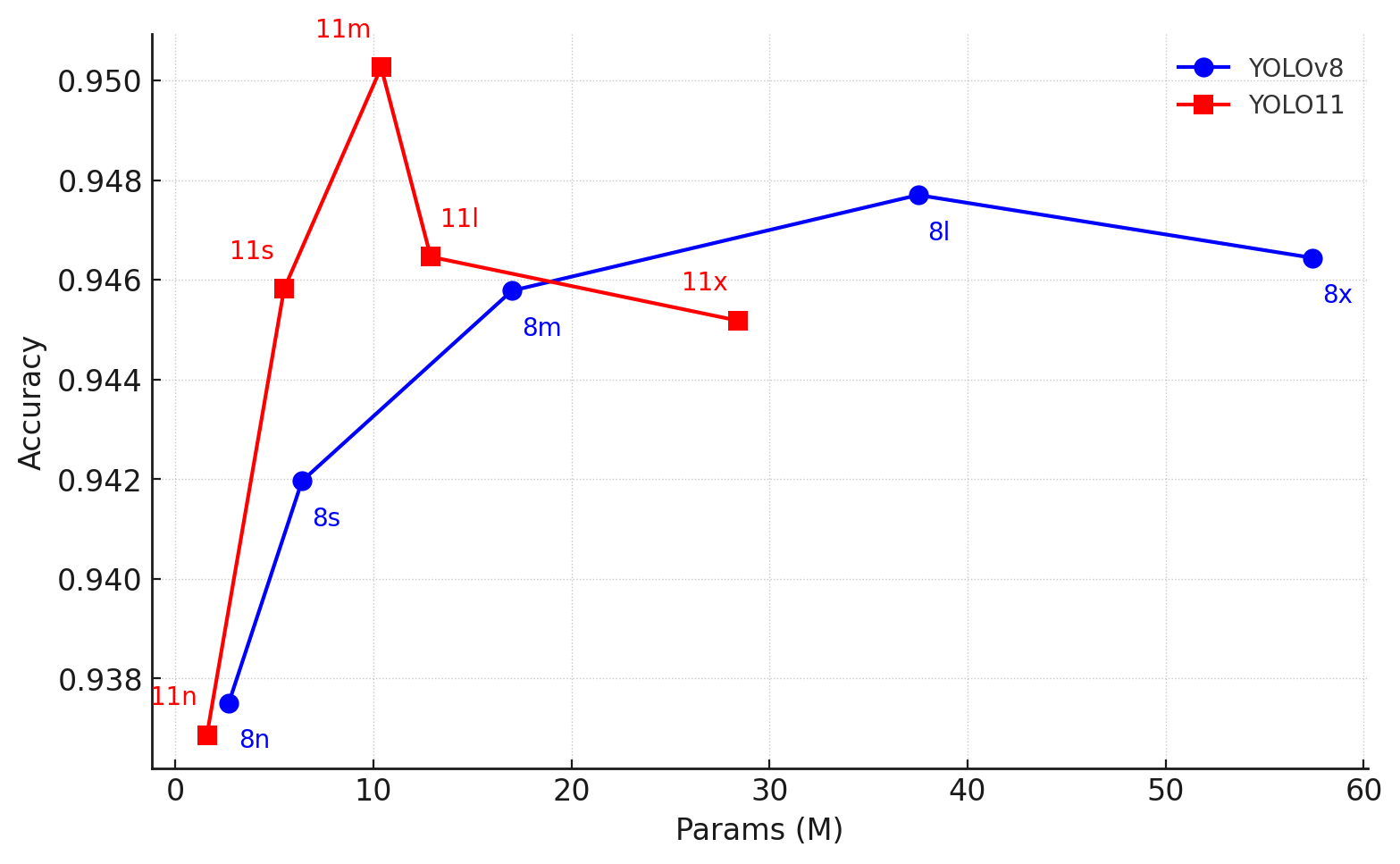}
\caption{Comparison of YOLOv8 vs YOLOv11 classification models across five size variants (nano (n), small (s), medium (m), large (l), and extra-large (x)), showing model parameter count (x-axis, millions) versus 5-fold cross-validation mean accuracy (y-axis).YOLOv8 variants are shown as blue circular markers connected by solid lines; YOLOv11 variants as red square markers connected by solid lines. Each marker is labeled by its size abbreviation.}
\label{figure4}
\end{figure}

We first compared the classification performance of YOLOv8 and YOLO11 across five model sizes (nano to extra-large). Under augmented training, YOLO11m-cls achieved the highest overall accuracy (95.03\%) and was therefore selected for further comparison. Compared to YOLO11s (94.58\%), YOLO11l (94.64\%), and all YOLOv8 variants (e.g., YOLOv8l: 94.77\%), YOLO11m-cls demonstrated the best balance between model size (10.4 M), latency (4.5 ms), and predictive performance (Table ~\ref{tab:yolo_comparison_percent}; Fig.~\ref{figure4}).YOLO11m-cls further improves upon YOLOv8 by introducing C3k2 blocks for efficient convolution, C2PSA attention modules for enhanced spatial focus, and a streamlined neck for better feature calibration. These additions collectively enhance its ability to localize and classify structurally complex or partially overlapping cell groups, while maintaining a low parameter count (10.4 M) and fast inference speed (4.5 ms).


We further benchmarked YOLO11m-cls against a wide range of classical convolutional and transformer models under the same experimental protocol. Without augmentation(Table ~\ref{tab:aug_comparison}), YOLO11m-cls already outperformed all competing models, achieving 92.47 ± 1.22\% accuracy and an F1-score of 93.19 ± 1.15\%—outperforming DenseNet121, the strongest baseline, by nearly 2 points. When trained with augmentation(Table ~\ref{tab:aug_comparison}), its performance improved further to 95.03 ± 0.77\% accuracy and 95.53 ± 0.65\% F1-score, maintaining a consistent margin over the best CNN-based model (VGG16). These findings underscore YOLO11m-cls’s ability to effectively capture both global and local structure, adapt to augmented data, and maintain efficiency—making it ideally suited for real-time classification of circulating cell clusters in microscopy-based imaging flow cytometry.

The superior performance of YOLO11m-cls can be attributed to several architectural advantages. Unlike traditional CNNs that rely on fixed-resolution features extracted from the final layers, YOLO-based classifiers adopt a detection-style architecture that incorporates multi-scale feature aggregation through PAN and SPPF, as well as deep semantic encoding in the backbone. This enables the model to retain both high-resolution spatial cues and rich contextual information—critical for distinguishing subtle morphological differences between single cells and tightly packed cell clusters.

Given these characteristics, YOLO11m-cls is particularly well-suited for circulating cell cluster (CCC) classification tasks in microscopy-based imaging flow cytometry, where subtle cell boundary variations and overlapping fluorescence signals make the problem inherently challenging. Its strong generalization across both augmented and non-augmented settings further supports its robustness and practicality in real-world biomedical screening scenarios.

\begin{table}[htbp]
\centering
\caption{Comparison of YOLOv8 and YOLO11 classification models on our dataset. The best accuracy and lowest latency are bolded.}
\label{tab:yolo_comparison_percent}
\begin{tabular}{lrrr}
\toprule
Model & Mean Accuracy (\%) & Mean Latency (ms) & Params (M) \\
\midrule
YOLOv8n-cls  & 93.75 & \textbf{3.0} & 2.7 \\
YOLOv8s-cls  & 94.20 & 3.1 & 6.4 \\
YOLOv8m-cls  & 94.58 & 3.9 & 17.0 \\
YOLOv8l-cls  & 94.77 & 4.8 & 37.5 \\
YOLOv8x-cls  & 94.64 & 5.2 & 57.4 \\
YOLO11n-cls & 93.69 & 3.8 & 1.6 \\
YOLO11s-cls & 94.58 & 3.9 & 5.5 \\
YOLO11m-cls & \textbf{95.03} & 4.5 & 10.4 \\
YOLO11l-cls & 94.64 & 6.2 & 12.9 \\
YOLO11x-cls & 94.52 & 6.6 & 28.4 \\
\bottomrule
\end{tabular}
\end{table}

\begin{table}[htbp]
\centering
\caption{Performance of different models under Non-Augmentation and Augmentation settings.}
\label{tab:aug_comparison}
\begin{adjustbox}{max width=1.1\textwidth,center}
\begin{tabular}{llcccc}
\toprule
 & Model & Accuracy & Precision & Recall & F1-score \\
\midrule
\multirow{8}{*}{\textbf{Non-Aug}} 
& AlexNet~\cite{krizhevsky2012imagenet}      & 88.20\% ± 1.08 & 88.30\% ± 1.16 & 87.92\% ± 1.02 & 88.05\% ± 1.08 \\
& DenseNet~\cite{huang2017densely}           & 90.56\% ± 2.06 & 90.66\% ± 2.04 & 90.35\% ± 2.14 & 90.45\% ± 2.10 \\
& EfficientNet~\cite{tan2019efficientnet}    & 80.67\% ± 2.97 & 80.77\% ± 2.59 & 80.29\% ± 3.28 & 80.36\% ± 3.23 \\
& ResNet18~\cite{he2016deep}                 & 86.04\% ± 1.61 & 86.09\% ± 1.48 & 85.83\% ± 1.74 & 85.88\% ± 1.67 \\
& ResNet34~\cite{he2016deep}                 & 86.42\% ± 1.80 & 86.53\% ± 1.60 & 86.15\% ± 2.01 & 86.24\% ± 1.90 \\
& VGGNet~\cite{simonyan2014very}             & 88.90\% ± 1.43 & 88.84\% ± 1.47 & 88.85\% ± 1.40 & 88.82\% ± 1.43 \\
& ViT~\cite{dosovitskiy2020image}            & 76.28\% ± 4.06 & 76.33\% ± 4.17 & 75.87\% ± 3.84 & 75.95\% ± 3.96 \\
& YOLO11m-cls~\cite{jocher2023yolov8}        & \textbf{92.47\% ± 1.22} & \textbf{91.99\% ± 1.88} & \textbf{94.50\% ± 2.63} & \textbf{93.19\% ± 1.15} \\
\midrule
\multirow{8}{*}{\textbf{Aug}} 
& AlexNet~\cite{krizhevsky2012imagenet}      & 93.17\% ± 1.33 & 93.26\% ± 1.44 & 93.01\% ± 1.30 & 93.10\% ± 1.34 \\
& DenseNet~\cite{huang2017densely}           & 94.26\% ± 1.26 & 94.34\% ± 1.24 & 94.11\% ± 1.31 & 94.20\% ± 1.28 \\
& EfficientNet~\cite{tan2019efficientnet}    & 93.94\% ± 1.68 & 94.01\% ± 1.77 & 93.80\% ± 1.67 & 93.88\% ± 1.69 \\
& ResNet18~\cite{he2016deep}                 & 93.94\% ± 1.28 & 93.98\% ± 1.26 & 93.93\% ± 1.38 & 93.89\% ± 1.30 \\
& ResNet34~\cite{he2016deep}                 & 94.45\% ± 1.61 & 94.47\% ± 1.66 & 94.36\% ± 1.59 & 94.40\% ± 1.62 \\
& VGGNet~\cite{simonyan2014very}             & 94.32\% ± 1.67 & 94.48\% ± 1.80 & 94.12\% ± 1.62 & 94.26\% ± 1.68 \\
& ViT~\cite{dosovitskiy2020image}            & 81.25\% ± 2.78 & 81.30\% ± 2.92 & 81.04\% ± 2.57 & 81.06\% ± 2.71 \\
& YOLO11m-cls~\cite{jocher2023yolov8}        & \textbf{95.03\% ± 0.77} & \textbf{96.26\% ± 0.67} & \textbf{94.71\% ± 1.44} & \textbf{95.53\% ± 0.65} \\
\bottomrule
\end{tabular}
\end{adjustbox}
\end{table}

\begin{figure}
\centering
\includegraphics[width=1.0\linewidth]{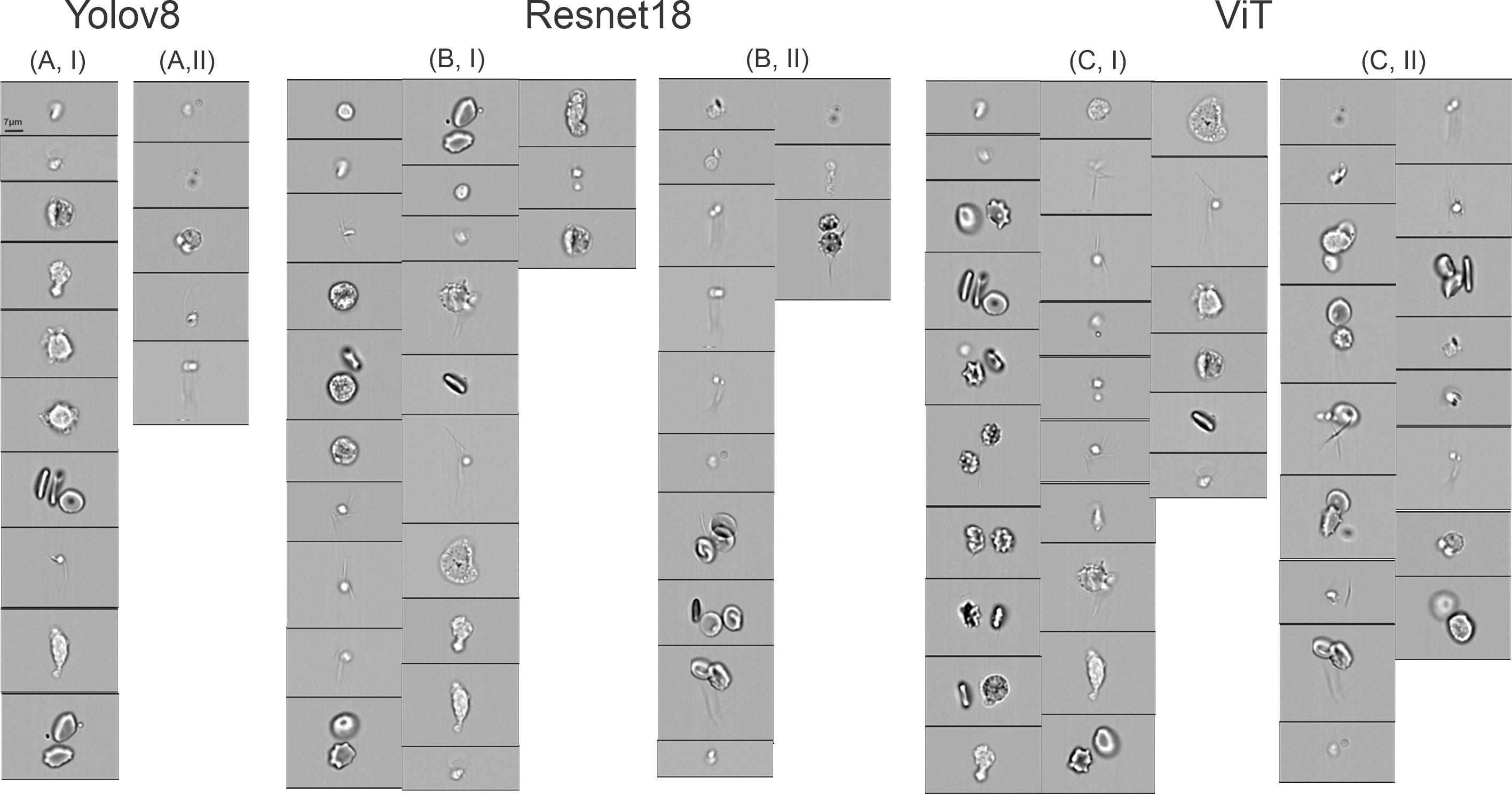}
\caption{Examples of misclassified images by (A) YOLO11, (B) ResNet18 and (C) ViT. (A, B, C, I) display the images containing no cell clusters are misclassified as cell clusters.   (A, B, C, II) display the images containing cell clusters are misclassified as non-cluster images. }
\label{figure5}
\end{figure}

In comparison, we showed the misclassified images of 2 typical classification models, ViT and Resnet18 model. Vit model shows a lower mean accuracy of 81.25\%, with a higher number of classification errors—17 in the cluster category and 26 in the non-cluster category, as illustrated in Fig.~\ref{figure5}C (I-II), respectively. These results show that the ViT model underperforms, particularly in cases where larger, morphologically distinct cells are present in the images, and they are erroneously classified as clusters despite being clearly separated. Since ViT uses self-attention mechanisms designed to capture global context,  these mechanisms may fall short when dealing with local feature intricacies like cell boundaries and overlap. Moreover, ViT models require larger datasets for effective training, and the limited size of the dataset applied here could constrain their capability, leading to performance that lags behind ResNet18 and YOLOv11. On the other hand, YOLOv11 exhibits superior performance, with relatively fewer errors, suggesting that it successfully learned the intricate features necessary for accurately distinguishing between the images containing clustered and non-clustered cells, such as the distance between the cells. 


\subsection{Results of classifying cell cluster phenotype.}

As introduced in the method section, we classify the phenotype of CCCs into RBC clusters, platelet clusters, WBC clusters, and WBC+platelet clusters by assessing the overlapping between cell cluster masks from the brightfield and the masks from the To evaluate instance segmentation performance, we selected a representative set of YOLO-based models that are officially released and publicly accessible with pretrained weights: YOLOv8, YOLOv9, and YOLO11. These models were chosen based on their architectural maturity, wide community adoption, and compatibility with our annotation and training pipeline.

We constructed a segmentation dataset consisting of 372 cell cluster images manually annotated via SAM API. For evaluation, we adopted a 4:1 split ratio, using 80\% of the images for training and 20\% for testing. All models were trained under identical conditions and assessed on the same test set. Performance was measured using standard instance segmentation metrics: mask mAP@0.5:0.95, mask mAP@0.5, mask mAP@0.75, and box mAP@0.5:0.95.fluorescence-stained region. 

As shown in Table~\ref{tab:yolo_seg_results}, YOLOv8-seg consistently outperformed both YOLOv9 and YOLO11 across all evaluation metrics. Specifically, YOLOv8-seg achieved the highest overall segmentation performance with a mask mAP@0.5:0.95 of 85.61\%, followed by YOLOv9 (81.48\%) and YOLO11 (81.39\%). This superiority extends to both low and high IoU thresholds, with YOLOv8 scoring 92.81\% at mask mAP@0.5 and 92.38\% at mAP@0.75, indicating better precision in both coarse and fine-grained mask alignment. It also achieved the highest box-level mAP@0.5:0.95 (86.14\%), reflecting superior localization accuracy. Given this performance gap, we adopted YOLOv8-seg as the backbone for instance-level mask generation in our framework.

\begin{figure}[ht]
\centering
\includegraphics[width=1.0\linewidth]{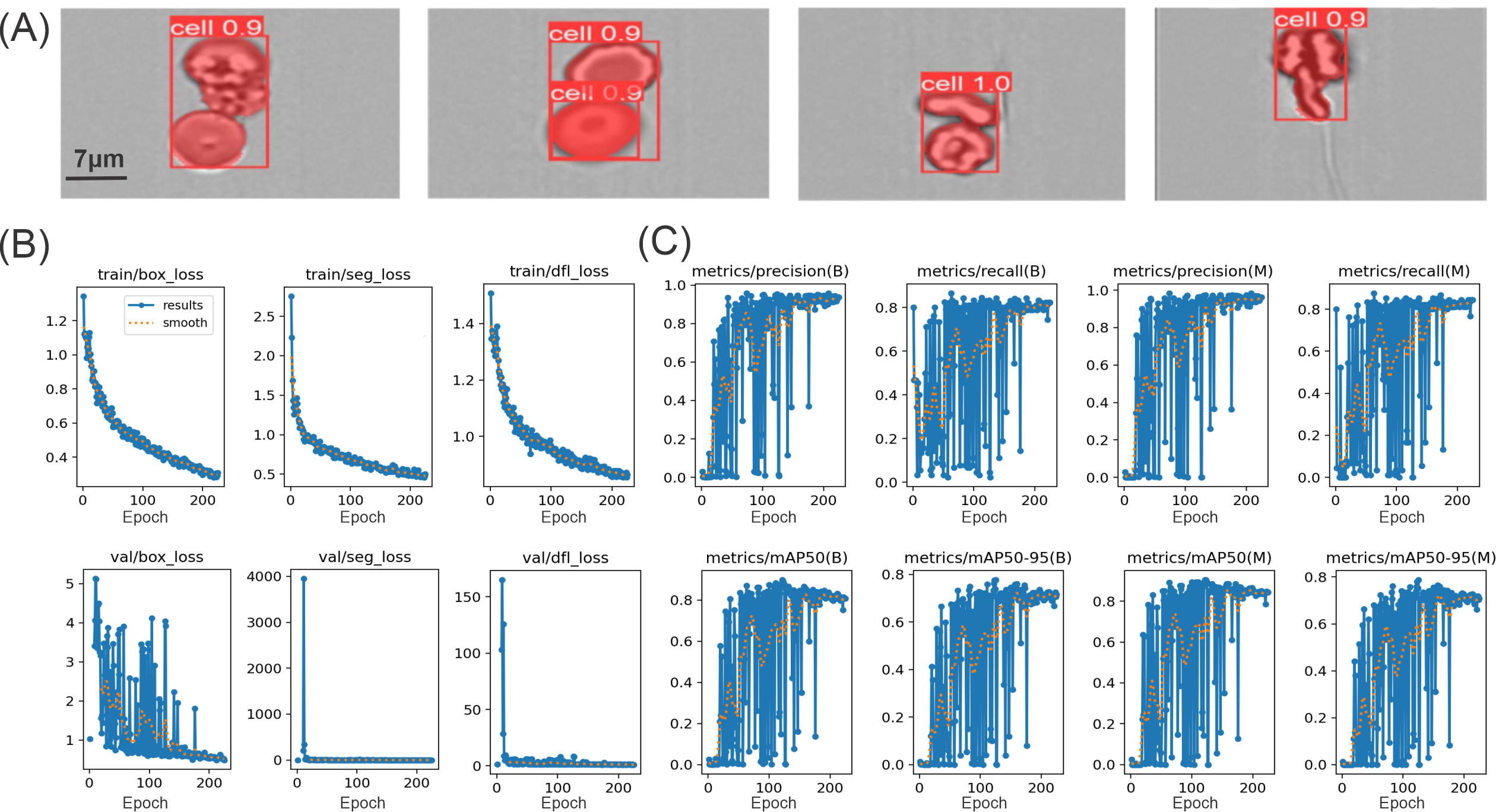}
\caption{
  Instance segmentation and training dynamics of the YOLOv8-seg model on cell cluster images. 
  (A) Representative segmentation results: red bounding boxes with confidence scores (e.g.\ ``cell 0.9'') and semi-transparent red masks overlay the detected cell instances. Scale bar, 7 $\mu$m. 
  (B) Training (top row) and validation (bottom row) loss curves over 200 epochs for bounding-box regression (\texttt{box\_loss}), mask segmentation (\texttt{seg\_loss}), and distribution focal loss (\texttt{dfl\_loss}); dotted lines show a smoothed trend. 
  (C) Epoch-wise evaluation metrics on the validation set for two cell classes (B, M): precision, recall, and mean average precision at IoU thresholds 0.5--0.95.
}
\label{figure6}
\end{figure}

Our results in Fig.~\ref{figure6}(A) show the YOLOv8-Seg model's capability to detect and segment cell clusters with high accuracy. Each detected cell is enclosed in a red bounding box with confidence scores of 0.9 or higher, indicating strong model confidence in this detection. The red masks show the segmentation of the cells, closely aligning with their boundaries. Fig.~\ref{figure6}(B) presents the training and validation losses for the YOLOv8 model during the process of cell cluster detection and segmentation. Fig.~\ref{figure6}(C) presents the precision, recall, and mean average precision (mAP) at different IoU thresholds (0.5 and 0.5-0.95) for both bounding box detection (metrics (B)) and segmentation (metrics (M)). These metrics show the model’s ability to accurately detect and segment cells. In summary, Figs.~\ref{figure6}(B-C) show a gradual improvement in the model’s performance, with decreasing losses and increasing precision, recall, and mAP metrics. The fluctuations in the validation metrics suggest some challenges in handling diverse cell structures, but the overall trends indicate successful training and enhanced detection and segmentation capabilities for cluster cell images.

\begin{table}[htbp]
\centering
\caption{Performance comparison of YOLO-based instance segmentation models (in \%).}
\label{tab:yolo_seg_results}
\begin{tabular}{lcccc}
\toprule
Model & mask mAP@0.5:0.95 & mask mAP@0.5 & mask mAP@0.75 & box mAP@0.5:0.95 \\
\midrule
YOLOv8  & \textbf{85.61} & \textbf{92.81} & \textbf{92.38} & \textbf{86.14} \\
YOLOv9  & 81.48 & 91.12 & 87.68 & 81.12 \\
YOLO11  & 81.39 & 91.34 & 85.11 & 81.58 \\
\bottomrule
\end{tabular}
\end{table}

\begin{figure}
\centering
\includegraphics[width=0.6\linewidth]{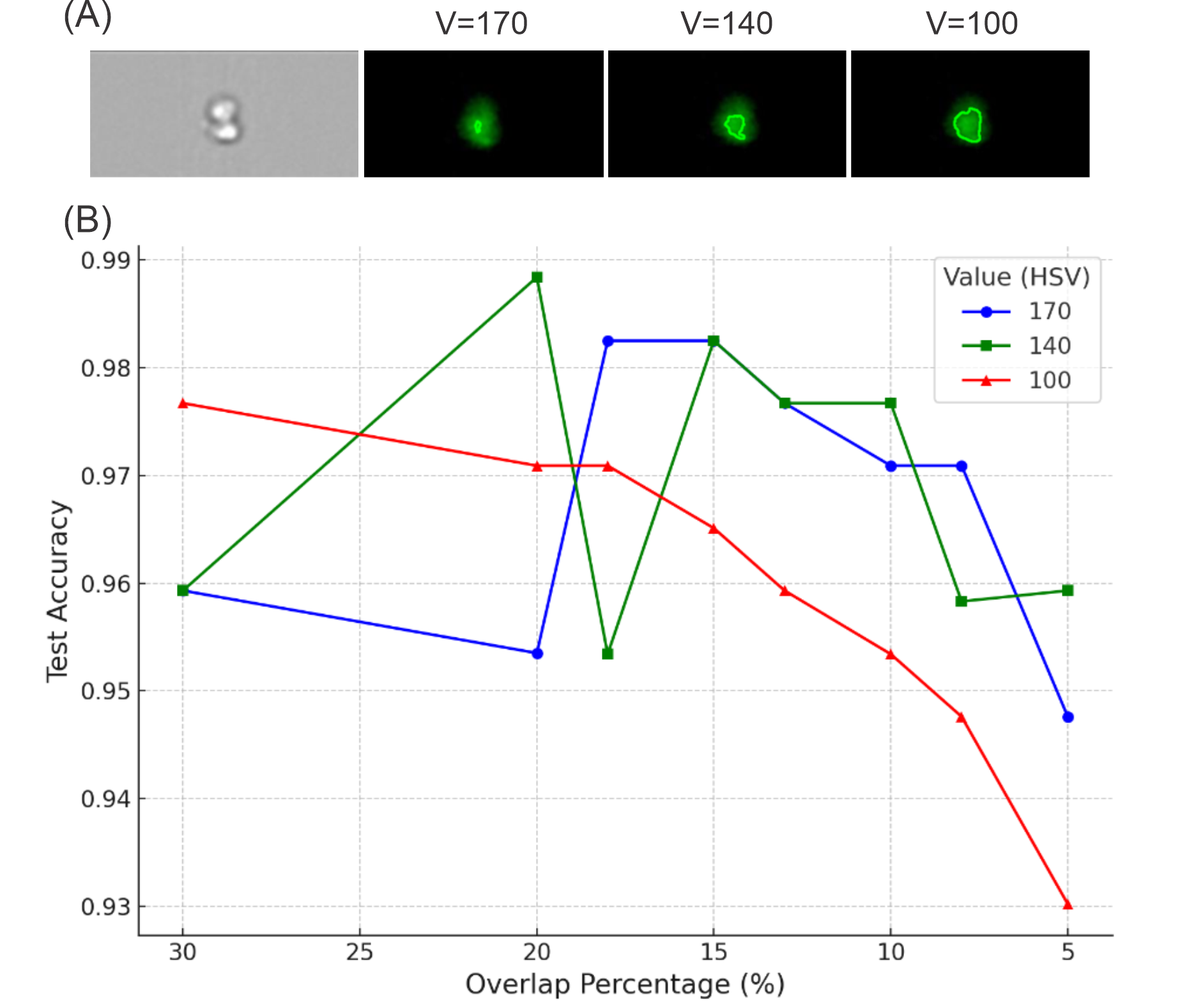}
\caption{(A) Impact of V value in HSV on detecting the contour of fluorescence-stained regions. (B) The impact of the overlap percentage between the  contour of fluorescence-stained regions and the contour of the cell cluster, as well as the V value on the prediction accuracy of the model.}
\label{figure7}
\end{figure}

In the process of analyzing cell cluster types, the determination of the specific types is heavily dictated by the overlapping between the fluorescence-stained cell regions and the original cell cluster regions, which serves as a key indicator of the cell cluster types. Given the complex morphology of cells and the variability in fluorescence staining intensity, it is not feasible to establish a universal brightness threshold that can accurately define the boundaries of the fluorescence-stained regions. To address this challenge, a series of experiments are conducted using three distinct brightness threshold values: 100, 140, and 170. Additionally, varying levels of percentage overlap are tested to identify the optimal combination of settings. As depicted in Fig.~\ref{figure7}(A), the fluorescence areas of interest are processed under the selected threshold values. These results indicate a direct relationship between the threshold value and the area of the fluorescence-stained region: higher threshold values corresponded to a decrease in the area of the stained region. Specifically, thresholds set above 170 results in fluorescence-stained regions that are too small, potentially compromising the accuracy of the results. Conversely, thresholds set below 100 resulted in excessively large regions that likely included noise unrelated to the actual cell clusters. Therefore, selecting an appropriate threshold range that effectively delineates the region of interest is a critical factor that directly impacts the outcome of the analysis.

The color of fluorescence-stained regions is used to identify specific cell cluster types, with green fluorescence-stained regions possibly corresponding to platelets and yellow fluorescence-stained regions potentially indicating WBCs. To accurately determine the presence of these specific cell types, we analyze the spatial relationship between the fluorescence-stained regions and the original cell clusters. The percentage of overlap between these regions is used as a key metric under the assumption that an overlap exceeding a certain threshold would be indicative of the presence of a particular cell type. However, due to the inherent complexity of cellular images and the potential for noise in the fluorescence staining, it is necessary to define a reasonable range for the overlap percentage. To this end, a series of gradient experiments are performed with overlap percentage thresholds set at 5\%, 8\%, 10\%, 13\%, 15\%, 18\%, 20\%, and 30\% to investigate their impact on the prediction accuracy. Fig.~\ref{figure7}(B) displays the dependence of test accuracy on the area overlap percentage thresholds and image brightness threshold values. This variability underscores the importance of selecting a threshold that provides consistent performance across different conditions. Our analysis reveals that when the overlap percentage threshold is set at 15\%, it produces relatively higher accuracy across all three brightness threshold values. This finding implies that a 15\% overlap threshold strikes a balance between sensitivity and specificity, effectively mitigating the impact of staining noise while still capturing the regions of interest. Furthermore, a brightness threshold value of 140 is identified as the most effective for defining the cell cluster regions.

\subsection{Visual Illustration }

For performance evaluation, we selected three representative classification models from different architectures: ResNet-18 (CNN-based), Vision Transformer (ViT; transformer-based), and YOLO11m-cls (from the YOLO series).  Gradient-weighted Class Activation Mapping (Grad-CAM)~\cite{selvaraju2017gradcam} was employed to visualize the regions that each model considers important for its predictions. Figure~\ref{figure7} illustrates the Grad-CAM heatmaps for two types of test images: (a) a non-clustered cell and (b) a clustered cell. For each case, the heatmaps generated by (I) ViT, (II) ResNet18, and (III) YOLO11m-cls reveal the models' attention to distinct features, providing insight into their classification strategies.

From the visualizations, it is evident that ViT (Figure A-I, B-I) focuses on specific, localized features, such as cell edges and internal structures, demonstrating its precision in isolating critical regions. ResNet18 (Figure A-II, B-II), in contrast, highlights broader regions, capturing both the central and peripheral areas of non-clustered cells and distributing attention across multiple cells in clustered scenarios. YOLO11m-cls (Figure A-III, B-III) concentrates its activations on sharp boundaries and overlapping regions, emphasizing spatial relationships and object localization, particularly in clustered cells.

Comparing the three models, ViT’s global attention mechanism enables the precise identification of key features, making it suitable for tasks requiring detailed classification. ResNet18’s hierarchical feature extraction balances local and global patterns, excelling in scenarios with complex structures. YOLO11, optimized for object detection, effectively handles boundary and spatial relationships. 

\begin{figure}
\centering
\includegraphics[width=0.6\linewidth]{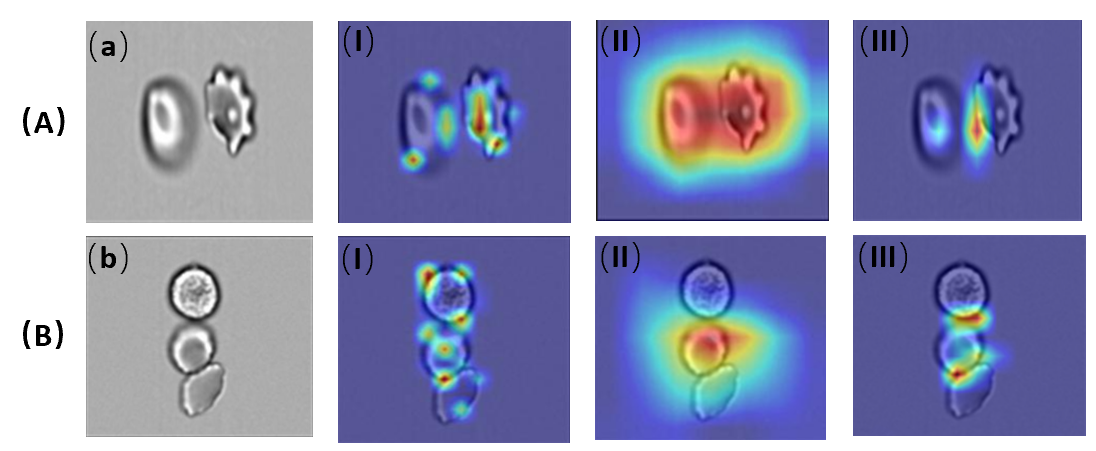}
\caption{Grad-CAM visualizations: (a) The Raw image of Non-cluster Cell, (b) The Raw image of Cluster Cell. Grad-CAM visualization of  (a) and (b) using the (I) Vit model, (II) Resnet18 model and (III) YOLO11 model respectively.}
\label{figure8}
\end{figure}

To evaluate the classification performance of the ViT, ResNet18, and YOLO11 models, Gradient-weighted Class Activation Mapping (Grad-CAM)~\cite{selvaraju2017gradcam} was used to visualize the regions that each model considers important for its predictions. Figure~\ref{figure8} illustrates the Grad-CAM heatmaps for two types of test images: (a) a non-clustered cell and (b) a clustered cell. For each case, the heatmaps generated by (I) ViT, (II) ResNet18, and (III) YOLO11m-cls reveal the models' attention to distinct features, providing insight into their classification strategies.

From the visualizations, it is evident that ViT (Figure A-I, B-I) focuses on specific, localized features, such as cell edges and internal structures, demonstrating its precision in isolating critical regions. ResNet18 (Figure A-II, B-II), in contrast, highlights broader regions, capturing both the central and peripheral areas of non-clustered cells and distributing attention across multiple cells in clustered scenarios.YOLO11m-cls (Figure A-III, B-III) concentrates its activations on sharp boundaries and overlapping regions, effectively emphasizing spatial relationships and object localization. This precise focus enables it to accurately differentiate overlapping cells and identify critical features, giving it a significant advantage in handling complex classification tasks. YOLO11’s superior performance can be attributed to its optimized architecture for object detection, which excels at identifying spatial hierarchies and boundaries, making it particularly well-suited for scenarios involving clustered or intricately structured cells.

\section{Discussion and Summary}

Circulating blood cell clusters and tumor cell clusters are important biomarkers for several pathological conditions, such as thrombotic complications~\cite{mackman2023circulating,phillips2015thrombotic,ward2021platelets} and those linked to inflammation~\cite{amintas2020circulating,gravsivc2023association}, autoimmune disorders~\cite{yasumizu2024single,walker2022link}, and viral infections like COVID-19~\cite{dorken2021circulating,dorken2023circulating,li2022multiphysics,mackman2023circulating,javadi2022circulating}  as well as cancer metastasis~\cite{hong2016circulating,schuster2021better,giuliano2018perspective}. The formation of cell clusters in these conditions and their numbers in the circulating blood could reflect underlying disease processes~\cite{aceto2014circulating,hong2016circulating}.  For example, one recent clinical study~\cite{dorken2021circulating} demonstrates the roles of circulating leukocyte clusters (CLCs), platelet-leukocyte aggregates (PLA), and platelet-erythrocyte aggregates (PEA) in the immunothrombotic state induced by COVID-19. Specifically, patients with COVID-19 had significantly higher levels of PEAs and PLAs compared to healthy controls. Among COVID-19 patients, CLCs were correlated with thrombotic complications, acute kidney injury, and pneumonia, whereas PEAs were associated with positive bacterial cultures. Thus, analyzing these cell clusters provides valuable insights into the pathogenesis of the diseases and helps clinicians to make disease prognosis. Flow cytometry, especially when paired with fluorescence staining~\cite{mckinnon2018flow,veal2000fluorescence}, is a powerful technique for examining cell clusters within blood samples. This method not only captures detailed information on cell morphology but also reveals protein expression profiles, allowing for precise identification of the cell types within the clusters.

In traditional flow cytometry analysis, manual gating is commonly used to identify and classify different cell populations. This process involves manually selecting regions on scatter plots or histograms to define cell types based on fluorescence intensity or other parameters. However, manual gating has several limitations, particularly in terms of efficiency and objectivity. In the last decade,  machine learning and deep learning methods have made significant progress in automating the analysis of single-cell flow cytometry images~\cite{verschoor2015introduction,rahim2018high,lee2019high,meehan2014autogate,malek2014flowdensity,conrad2019implementation}. However, there has been limited effort in developing tools to analyze images containing complex cell clusters. In contrast to single cells, cell clusters typically display irregular shapes and sizes. Additionally, these clusters often contain a mix of different cell types, necessitating multi-channel staining to accurately identify the specific cells within the clusters. To fill this methodology gap, we present a new computational framework specifically designed for the automatic classification of circulating cell clusters. This framework can effectively identify the different cell types within clusters, overcoming the challenges posed by the irregular shapes and sizes that distinguish cell clusters from single cells.

Since these clusters are often composed of a mix of heterogeneous cell types, multi-channel fluorescence staining is required to properly identify the specific cells involved. The classification of multi-channel flow cytometry images of blood cell clusters presented several challenges, including varied boundaries among different types of cells, overlapping of cells within the clusters, and the presence of irregular noise artifacts, all of which complicated the classification task. Based on these challenges, our framework uses a two-step analysis strategy. First, we classify the images into cell cluster and non-cell cluster groups using a modified YOLO11 model. This model has demonstrated superior performance when compared to the CNN series and the Vit models for this task. In the second step, we determine the cell types within each cluster by matching the contours of the clusters with those extracted from multiple fluorescence staining channels. This approach allows for the accurate classification of cell cluster phenotypes while avoiding interference from cell debris and staining artifacts. Our results show that the proposed computational framework can achieve over 95\% accuracy for both cluster and non-cluster classification as well as for phenotype identification within cell clusters, demonstrating the framework's effectiveness.

While initially developed for circulating blood cells, our computational framework can be extended to analyze other cell cluster types, including circulating tumor cell (CTC) clusters~\cite{liu2021simultaneous,liu2024ctc,chaffer2011perspective,lambert2017emerging}. CTCs play a key role in cancer metastasis, primarily existing as single cells, yet recent studies reveal that CTC clusters significantly enhance both survival rates and metastatic efficiency~\cite{aceto2014circulating,fabisiewicz2017ctc,hong2016circulating}. These clusters are composed of CTCs, immune cells, platelets, and extracellular matrix components. A primary challenge in identifying CTCs and CTC clusters is the heterogeneity in CTC morphology and biomarker expression (e.g., CK, EpCAM, Vimentin)~\cite{lin2021circulating,zhao2019tumor}. The precision of our model in analyzing bright-field and fluorescence-stained images creates a pathway for identifying CTC clusters. The identified cell clusters can also be used to set up computational models~\cite{li2022multiphysics,li2024red,li2022computational,li2023combined,javadi2022circulating,yazdani2021integrating} to simulate the transport of CCCs or CTC within the microcirculation and their potential of blocking blood flow or tumor extravasation.

The proposed framework holds potential for clinical translation in hematological and immunological diagnostics. By enabling automated, real-time identification and phenotyping of circulating cell clusters (CCCs), the system could assist in early detection of thrombosis-prone states, systemic inflammation, or aberrant immune cell aggregation, especially in COVID-19, sepsis, or cancer. Its compatibility with standard flow cytometry imaging platforms makes it a promising candidate for integration into clinical pipelines, potentially reducing manual gating bias and increasing diagnostic throughput.

In summary, we have developed a computational framework that automatically and accurately analyzes flow cytometry images of CCCs using both bright-field and fluorescence-stained images. We have validated the framework’s effectiveness by benchmarking its performance against peer models. This framework opens up possibilities for cellular-level research across various human diseases, facilitating better understanding and diagnosis.

\subsection{Limitations and Future Directions}

Despite the strong performance achieved by our two-stage classification and phenotyping framework, several limitations remain.

First, the current phenotyping relies on threshold-based HSV segmentation and a fixed overlap ratio criterion, which may be sensitive to illumination variability and staining inconsistencies. Future work may explore integrating learned feature-level fusion strategies (e.g., cross-channel attention) or weakly supervised multi-label learning for more robust phenotypic classification. Second, although the YOLOv8-seg and YOLO11m-cls architectures demonstrated superior accuracy and speed trade-offs, the segmentation and classification modules are trained separately. An end-to-end, jointly optimized pipeline could further improve performance and inference efficiency by sharing features across tasks. Lastly, while our study focuses on COVID-19 patient samples, circulating cell clusters are clinically relevant across a broader range of diseases (e.g., cancer, thrombosis). Future research should explore transferability and domain adaptation techniques to generalize the framework across diverse pathological contexts.

\section*{Funding}
This work was supported by National Institute of Health grant R21HL168507 and NSF SCH Award Number: 2406212. M.X. is partially supported by the DOE SEA-CROGS project (DE-SC0023191) and AFOSR project (FA9550-24-1-0231). High-performance computing resources were provided by The Georgia Advanced Computing Resource Center (GACRC) at the University of Georgia.

\section*{Acknowledgments}
H.L. would like to thank Shan-Ho Tsai from the Georgia Advanced Computing Resource Center (GACRC) for providing technical support.

\bibliographystyle{unsrt}
\bibliography{reference}

\end{document}